\documentclass[sigconf,nonacm]{acmart}

\usepackage{algorithm}
\usepackage{algorithmic}
\usepackage{subfigure}

\def\BibTeX{{\rm B\kern-.05em{\sc i\kern-.025em b}\kern-.08emT\kern-.1667em\lower.7ex\hbox{E}\kern-.125emX}}
    
\begin{document}

%
\title{A Generalized Framework for Population Based Training}

%
\author{Ang Li}\email{anglili@google.com}
\affiliation{%
  \institution{DeepMind, Mountain View}
}

\author{Ola Spyra}
\email{aspy@google.com}
\affiliation{%
  \institution{DeepMind, Mountain View}
}

\author{Sagi Perel}
\email{sagipe@google.com}
\affiliation{%
  \institution{Google Brain, Pittsburgh}
}

\author{Valentin Dalibard}
\email{vdalibard@google.com}
\affiliation{%
  \institution{DeepMind, London}
}
\author{Max Jaderberg}
\email{jaderberg@google.com}
\affiliation{%
  \institution{DeepMind, London}
}
\author{Chenjie Gu}
\email{gcj@google.com}
\affiliation{%
  \institution{DeepMind, Mountain View}
}

\author{David Budden}
\email{budden@google.com}
\affiliation{%
  \institution{DeepMind, London}
}

\author{Tim Harley}
\email{tharley@google.com}
\affiliation{%
  \institution{DeepMind, London}
}

\author{Pramod Gupta}
\email{pramodg@google.com}
\affiliation{%
  \institution{DeepMind, Mountain View}
}

%
\renewcommand{\shortauthors}{Li et al.}

%
\begin{abstract}
Population Based Training (PBT) is a recent approach that jointly optimizes neural network weights and hyperparameters which periodically copies weights of the best performers and mutates hyperparameters during training. Previous PBT implementations have been synchronized glass-box systems. We propose a general, black-box PBT framework that distributes many asynchronous \linebreak ``trials'' (a small number of training steps with warm-starting) across a cluster, coordinated by the PBT controller. The black-box design does not make assumptions on model architectures, loss functions or training procedures. Our system supports dynamic hyperparameter schedules to optimize both differentiable and non-differentiable metrics. We apply our system to train a state-of-the-art WaveNet generative model for human voice synthesis. We show that our PBT system achieves better accuracy and faster convergence compared to existing methods, given the same computational resource.
\end{abstract}

%
%


%

%

%
\maketitle

\section{Introduction}
Neural network training typically employs a two-stage procedure, \textit{i.e.}, hyperparameter tuning followed by model training. Although there have been systems that enable automatic optimization of the hyperparameters, the gap between the two stages relies heavily on human engineering which results in a lengthy and often inefficient model development cycle. Deep neural networks are especially powerful with large scale data, however the labor costs in model training and tuning on large datasets are becoming more expensive at the same time. Nowadays, machine learning professionals spend a majority of their time tuning model-related parameters. Effectively automating the neural network training procedure becomes an important yet challenging problem to address. A promising direction is to build an end-to-end machine learning system that merges the two-stage learning process and reduces human engineering.

Population Based Training (PBT) has recently emerged and opened up a new direction in neural network training which jointly optimizes the hyperparameters while training the network weights at the same time \cite{max-pbt-2017}. The core idea is to repeatedly replace a poorly performing model with a better performer and continue training with hyperparameters mutated from the better one. The mutation is an important procedure that allows the hyperparameters to dynamically change over time, which is difficult to achieve by any conventional hyperparameter tuning method.

\begin{figure}[!tb]
\centering
\vskip 1em
\includegraphics[width=.475\textwidth]{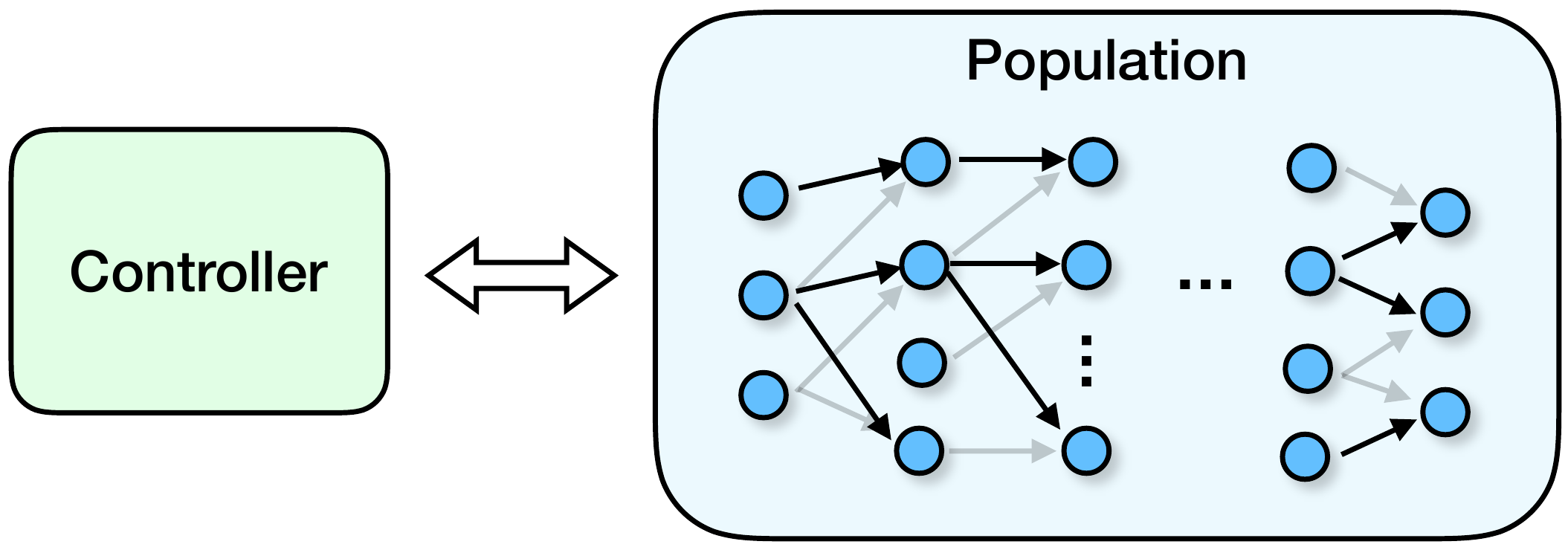}
\caption{Black-box Service for Population Based Training based on a Worker-Controller framework. Each solid blue circle represents a training trial. A black arrow represents a trial dependency (usually for warm-starting the model from a parent's checkpoint) and a gray arrow represents an un-selected parent trial which loses in a tournament and fails to reproduce. The PBT controller oversees the progress of the whole population and decides training actions.}
\label{fig:pull}
\end{figure}

A natural way to design a PBT system is to allow every worker access to the information of all other workers in a shared database and all of the workers progress at the same pace. However, this implementation approach typically requires continuous and simultaneous training of all workers and may encounter issues when the workers could be preempted by other higher priority jobs in a real distributed working environment. In addition, existing implementations of PBT are mostly glass-box approaches, which introduces additional restrictions on how the neural network model is implemented. The trainer has to know information about other parallel workers and performs weight copying and hyperparameter changes. The hyperparameters, for example, in TensorFlow \cite{tensorflow} may have to be defined in the computation graph in order to be changed efficiently every so often.

We propose a generalized Population Based Training framework to improve training extensibility and scalability. The proposed framework is based on the idea of decomposing the whole model training into multiple trials, where in each trial a worker only trains for a limited amount of time. Figure \ref{fig:pull} illustrates the high-level controller-worker framework adopted by the proposed system. An important notion of the system is that each trial is dependent on one or more other trials, \textit{e.g.}, the initial checkpoint can be the last checkpoint of another trial and the hyperparameters can be decided based on other trials' final measurements. A population controller is introduced into the system to oversee the whole population of trials. The controller also decides the hyperparameters and the checkpoint for warm-starting a worker in a new trial.

This paper introduces the system design of a black-box PBT service and evaluates the accuracy, performance and scalability of the proposed system by case study on the real world application of human speech synthesis using WaveNet \cite{wavenet}, where we show improved performance, in terms of both accuracy and convergence, compared to existing hyperparameter optimization methods \cite{vizier}.


\section{Preliminary}

The proposed PBT service framework is inspired by the design of Vizier hyperparameter tuning service \cite{vizier}. This section introduces the concepts of both Population Based Training and the Vizier hyperparameter tuning service.
\subsection{Vizier Service}
Vizier is a black-box hyperparameter optimization service developed by \citet{vizier}. Compared to other systems for hyperparameter tuning, the major advantage of Vizier is that a black-box service could significantly reduce the effort required for setting up a hyperparameter tuning experiment. In addition, a black-box service allows the highest flexibility in the setup of the training procedure in the client side, \textit{i.e.}, it can be easily applied to different types of machine learning models and model training libraries.

\subsection{Population Based Training}
Population Based Training (PBT) was proposed by \citet{max-pbt-2017}; it is an asynchronous optimization algorithm that effectively utilizes a fixed amount of computational budget to jointly optimize a population of models and their hyperparameters. PBT is related to evolutionary strategies, however it differs from conventional evolution in that PBT employs an idea called warm-starting, \textit{i.e.}, initializing a model training session using a checkpoint saved from another model's training.

The reason why PBT performs efficient hyperparameter search is because PBT makes decisions based on incomplete observations (non-converged objective values). Most traditional hyperparameter tuning methods require training until near convergence and use the final performance measurement to guide the following search. Those processes could be extremely lengthy especially in the large scale deep learning scenarios.

\textit{Glass-Box Implementation.} All of the existing implementations of PBT are glass-box approaches where the parallel workers read and write to the same database and decide whether to warm start from another worker's checkpoint. The glass-box implementations have following limitations:
\begin{itemize}
\itemsep 0pt
    \item Any changes made to the computation graph can be complicated. So the hyperparameters often need to be defined in the computation graph.
    \item In a distributed setting, the glass-box approach does not gracefully handle the case of a worker job being preempted by another worker job.
    \item The glass-box framework is not extendable to advanced evolution or mutation decisions that need to be made based on a global assessment of all the workers' performance measures.
\end{itemize}

We propose a black-box service based solution to PBT training. We show that our black-box design can address the above issues encountered by the glass-box approaches.

\section{PBT Service Framework}

\subsection{Overview}
\begin{figure}[!tb]
    \centering
    \includegraphics[width=\linewidth]{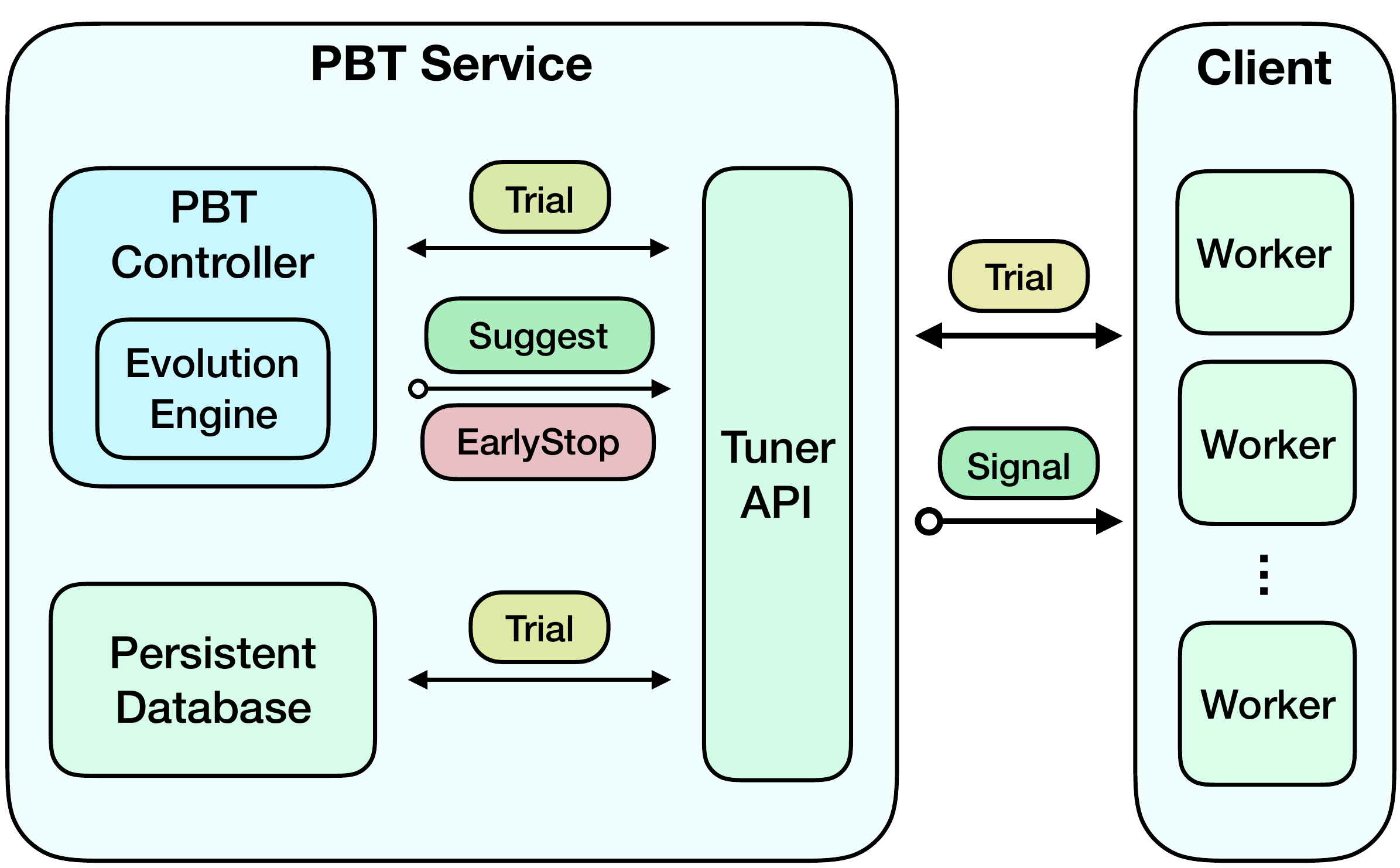}
    \caption{PBT Service System Diagram. The PBT service is composed of a controller, a tuner API layer and a persistent database. Tuner API and the database are similar to the design of the Vizier service. The major information container about model training is called \textit{Trial}, defined using a protocol buffer, which is passed between PBT controller, Tuner API, persistent database and the workers in the client. PBT controller may suggest two kinds of actions to the client, \textit{i.e.}, suggest a new trial or early-stop a trial.}
    \label{fig:pbt-system}
\end{figure}

The proposed PBT service is a \emph{stateless} service, by which we mean each of the requests to the service does not depend on any other. This generally follows the design of the Vizier hyperparameter tuning service and allows the service to be highly scalable. Figure \ref{fig:pbt-system} shows a system diagram of the proposed PBT service.

The PBT service is a black-box model training service which jointly optimizes hyperparameters and model parameters. Instead of asking a worker to make decisions on its own, we turn over all the decision making to a central controller. The workload assigned to each trial becomes a relatively smaller number of training steps and the workers always send requests to the controller (or service) for their next moves. Our PBT service has the following advantages:
\begin{itemize}
\itemsep 0pt
    \item Allows for tuning hyperparameters no matter whether they are defined in the computation graph or not.
    \item Allows for training a model with both differentiable and non-differentiable objectives.
    \item Allows all hyperparameters to be dynamic over time.
    \item Sufficient scalability and flexibility in using low priority workers.
    \item Maximizes flexibility with the machine learning model training frameworks.
\end{itemize}

We introduce the concepts and major components of the proposed PBT service in the rest of this section.

\subsection{Trial}
A \textit{trial} represents a continuous training session. The configuration of a trial is defined using a protocol buffer (protobuf), which is a critical piece of information exchanged frequently among PBT controller, tuner API and the workers.

The main fields in the trial protobuf are parameters and metadata where metadata contains information that should not be regarded as hyperparameters. Parameters contain all the hyperparameters of a trial and the metadata stores PBT algorithm related messages. So a trial typically contains the following fields:
\begin{itemize}
\itemsep 0pt
\item \textit{hparams}: The hyperparameters of the training session.
    \item \textit{warm\_start\_checkpoint\_path}: An optional field indicating which checkpoint path the training session should start with.
    \item \textit{parent\_trial\_id}: An optional field indicating the trial ID of the current trial's parent (where the warm start checkpoint comes from).
    \item \textit{initiator\_parent\_trial\_id}: An optional field indicating the trial that initiated the reproduction that results in the current trial. The initiator may not be the real parent of the current trial.
\end{itemize}

\subsection{Parameters}
The PBT system supports four types of parameters: integer, floating point, discrete and categorical values. A user should specify flexible hyperparameters at the beginning of a training study. They could also disable the evolution (or mutation) of some parameters. The mutator in the evolution engine refers to this field and skips the mutation of any parameters specified in it.

It is possible some parameters are dependent on other parameters, \textit{e.g.}, learning rate is dependent on the type of optimizer. So essentially, all the parameters are represented using a directed acyclic graph (DAG). 

\subsection{Controller}
We follow the algorithm playground design in Vizier and define the following two functions for the population controller, which are elaborated below.
\begin{itemize}
\itemsep 0pt
    \item \textsc{GetNewSuggestion}(\textit{trials}, $k$): Return a list of $k$ new trials given all the existing trials. The trials returned from this method are basically determined by the reproduction procedure in the evolution engine. Please see the next section for more details.
    \item \textsc{GetEarlyStoppingTrials}(\textit{trials}): Return a list of trials that need to be stopped immediately given all existing trials. This function is useful when we want to kill some running trials and free the resources for other trials. Although not a necessary component, it allows for a more generalized evolution framework.
\end{itemize}

\subsection{Initiator Based Evolution}
Evolution algorithms have been extensively studied over the past decades. While there are numerous strategies in the evolution literature, we introduce an initiator based evolution framework. The framework is well aligned with the original PBT algorithm \cite{max-pbt-2017} which implements the explore/exploit operations. Our proposed framework is generic and can be extended to many other evolution approaches. A major advantage of such initiator based approach is that every trial is guaranteed to participate in at least one reproduction procedure. We believe such guarantee is important for evolution with a small population size such as 20. We follow the popular evolution design choice by formulating the fitness of each trial, and defining the reproduction strategy.


\subsubsection{Fitness representation}
The fitness of a trial represents how well a trial performs. A higher fitness leads to higher chance of survival in evolution. We represent the fitness of a trial as a tuple
\begin{equation}
    f(\mathbf t) = [f_1(\mathbf t), f_2(\mathbf t), \ldots, f_k(\mathbf t)]
\end{equation}
where $\mathbf t$ is a trial and $f_i$ is the $i$-th fitness evaluation function. 

The reason for defining such generalized tuple representation is because many real world applications contain multiple objectives and it is often the case that some objectives (fitnesses) are missing due to noisy or missing data. Representing all objectives into the fitness allows a thorough and robust comparison between trials.

The comparison of two fitness tuples can be defined in different ways for different applications. For multi-objective optimization, a fitness $f_a$ is better than $f_b$ if and only if all elements in $f_a$ are larger than their correspondences in $f_b$. However, the fitness can also be compared with priority, \textit{e.g.}, first compare the first element and then compare the rest only if the first elements are equivalent.



\subsubsection{Reproduction Strategy}
The main reproduction concepts in the proposed initiator based evolution is described as follows.

\paragraph{Initiator.} We define the concept of an \textit{initiator}, which represents a trial that initiates the current reproduction. This is an asymmetric evolution, \textit{i.e.}, when a trial completes, it sends a request to the server to initiate a competition with other population members to decide its next move.


\paragraph{Opponent Selection.} Not every member of the remaining population participates in the competition. We define a selection process which selects the population members in the last $k$ generation as the potential opponents. The value $k$ is empirically set default to 2.
However, it is often the case that the devices in a distributed setting are not homogeneous, \textit{i.e.}, different workers may perform differently. In addition, some parameters of the model may also affect the training and evaluation speed. To relieve such effect, we restrict every trial to only compete with trials in the past $k$ generations (including earlier generations and the same generation as the initiator trial). We justify this design choice through a comparative experiment, which will be described in Section \ref{sec:opponent}.

\paragraph{Parent.} The initiator competes against another trial randomly sampled from the survival pool. The winner is chosen as the parent for the current reproduction.
So essentially every population member participates in a binary tournament once and only once. This is also known as \textit{binary tournament selection}.

\paragraph{Reproduction.} A new trial (or child) is generated based on the selected parent through a procedure called \textit{reproduction}.
A typical reproduction procedure contains crossover (\textit{aka.} recombination) and mutation. However, the vanilla PBT algorithm is essentially a single parent evolution without a crossover. The mutation for scalar hyperparameters is to choose a random multiplier, either 0.8 or 1.2. In addition, we also implement mutations for categorical and discrete parameters. For categorical parameters, the mutation is equivalent to random sampling. For discrete parameters, we restrict the parameter to mutate to either the lowest larger element or the largest lower element \textit{w.r.t.} the current value.



\subsection{Worker}

\begin{figure}[!tb]
    \centering
    \includegraphics[width=.85\linewidth]{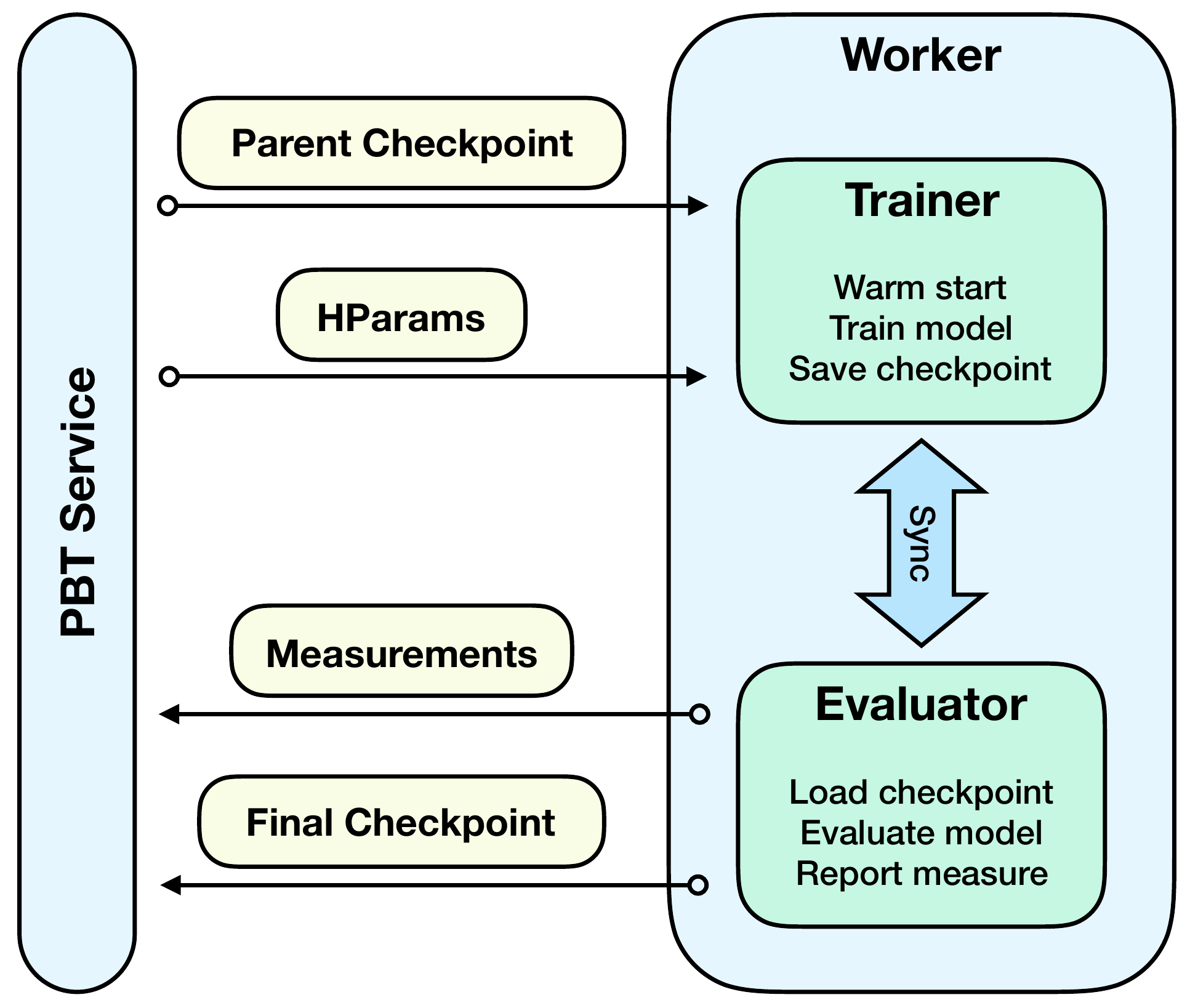}
    \caption{System diagram of a worker. A worker is composed of a trainer and an evaluator which are synchronized. The trainer receives messages such as warm start checkpoint path and hyperparameters from the PBT service and the evaluator sends messages about the measurements (objective values) and checkpoint paths back to the service.}
    \label{fig:worker}
\end{figure}

A worker represents one training process that is composed of a trainer and an evaluator. Figure \ref{fig:worker} shows an illustration of how the worker works. After a worker finishes its job, \textit{i.e.}, when the evaluator has evaluated the final checkpoint in the model directory, a worker will request a new trial suggestion from the PBT server.

\subsubsection{Trainer}
The trainer receives a parent checkpoint path and hyperparameters from the PBT service. The trainer constructs the model architecture using the given hyperparameters as if it were a vanilla training session. The parent checkpoint is restored into the corresponding variables when the model is constructed. This is why the PBT service design is a black-box approach -- the model training procedure does not require knowledge of PBT and the PBT system does not need to know about the internals of model training. The only addition to the trainer is the checkpoint warm starting before the training session starts.

\subsubsection{Warm-start.}
In usual cases, the warm-starting procedure can be done using checkpoint restoring such as \textit{tf.train.Saver.restore}\footnote{\url{https://www.tensorflow.org/guide/saved\_model}} in TensorFlow. However, there are cases where a hyperparameter affects the architecture of the neural network and a typical checkpoint restore method usually does not support. In those cases, we use a smart saver\footnote{A similar implementation example can be found in \url{https://github.com/yk/tensorflow-optimistic-restore-saver}} which analyzes the correspondence between a checkpoint and the model architecture using their variable names and only restores matched variables.

\subsubsection{Evaluator}
The evaluator conducts continuous evaluation of the checkpoints in the model directory of the corresponding trial.
Every checkpoint evaluation will report a measurement back to the PBT service, together with the checkpoint path. A final checkpoint has to be kept because it may serve for warm starting a future child trial.

\subsection{Garbage Collection}
A potential issue of large population based training is the exploding size of saved checkpoints. Although one training session can specify automatic garbage collection, as the total number of parallel workers increases, the total number of checkpoints can still be a significant number. We implement a global garbage collection as an option to work together with training jobs. It periodically reads all trial protobufs from the tuner's database, identifies all checkpoints that have already been evaluated and removes these checkpoints from the storage. The client can optionally keep the last checkpoint in every training job since they might be used for serving purposes.


\begin{algorithm}[!tb]
   \caption{PBT trial suggestion \textsc{GetNewSuggestion} function.}
   \label{alg:example}
\begin{algorithmic}
   \STATE {\bfseries Input:} population size $population\_size$, all completed and pending trials $trials$.
   \IF{ $\textsc{LastCompleteGeneration}(trials)==0$}
   \STATE $child=\textsc{SampleTrial}()$;
   \ELSE
   \STATE $initiator=\textsc{GetOldestUninitated}(trials)$;
   \STATE $opponents=\textsc{SelectOpponents}(initiator,trials)$;
    \STATE $child=\textsc{Reproduce}(initiator, opponents)$;
    \ENDIF
    \STATE \textbf{return} $child$;
\end{algorithmic}
\end{algorithm}

\subsection{Budget Mode}
The PBT algorithm requires a number of parallel workers to perform model training at the same time. However, it is often difficult to obtain sufficient devices available for a large number of workers at the same time. The proposed service also handles such cases by simulating the evolution with a large generation size using a small number of workers. This is called the \textit{budget} mode.

The budget mode is implemented in a straightforward way by always picking the oldest trial that has not yet initiated any reproduction and only starting a reproduction when the initiator's generation has reached the specified population size. Please note that the budget mode will also be a synchronized evolution when there is only one worker.




\subsection{Training Replay}


Population based training is more effective when using a large population size, which results in a large number of model snapshots with their hyperparameter trajectories. The best performing snapshot is often used for serving purpose. However, there is often a need to re-train the same model on a slightly different set of data. Re-training using a large population again and again may not be the optimal choice. We implement a feature called \textit{training replay}, which extracts and performs the same training procedure (trial dependency and hyperparameter trajectory).

\subsubsection{Subset Training Replay.} The PBT training replay also allows to replay a set of trials in any existing training study. This feature can be useful when a user wants to extract multiple models from the whole population. A particular use case is for model ensemble.

\subsubsection{Trial Dependency Graph.} All the trials are dependent on each other except the initial ones. The training replay requires to extract the dependency graph of a certain final trial in order to perform the same training procedure. This dependency graph is directed acyclic because all the trials are trained over time -- it is impossible for old trials to warm start from a new trial's checkpoint. So the dependency graph can be extracted by traversing the final trial node to an ancestor. A topological sort is performed to enforce the trial dependency into the execution order. 

\subsection{Training Recovery}
The PBT service is a stateless service which allows to seamlessly recover a paused or faulty training procedure. The status of all pending trials should be marked as stopped before resuming the PBT training. All of the trial information is passed into the controller as usual and the controller can return new trial suggestions for those trials that have not initiated a reproduction.


\section{Case Study: WaveNet}
While the effectiveness of Population Based Training has been demonstrated by \citet{max-pbt-2017} with a variety of applications such as neural machine translation, generative adversarial networks and reinforcement learning in DM-Lab, Starcraft, etc, we present in this paper a new application of PBT on speech synthesis using WaveNet, to conduct analysis on both accuracy and performance of the proposed PBT system.

\subsection{WaveNet for Human Voice Synthesis}
WaveNet is the state-of-the-art deep generative neural network in modeling raw audio waveforms \cite{wavenet}. The basic building block of a WaveNet is dilated causal convolution. A dilated convolution layer applies convolution with skipped input units which leads to an increased reception field of the convolution without any additional computation costs.

\subsubsection{Dataset.}
We evaluate our system using a public speech dataset, namely LibriSpeech \cite{Panayotov2015LibrispeechAA}. The dataset contains $1000$ hours of English reading speech and is split into training, validation and evaluation sets. We use the training and validation set for all model training.

\subsubsection{Setup.}
We train all models using Tesla P100 GPU cards. Each worker contains one trainer and one evaluator where the evaluator continuously evaluates the latest checkpoints in the trial's directory and reports the objective values to the PBT service. 
All trainers have 1 chief GPU worker, 16 GPU workers and 2 parameter servers. The model uses Adam optimizer with exponential moving average on the model weights. The only flexible parameter is learning rate. 

\subsection{Approaches}
We compare our PBT service against several popular hyperparameter tuning approaches on the LibriSpeech dataset. All approaches are summarized below:
\begin{itemize}
\itemsep 0pt
    \item \textit{Grid Search}: 5 parallel trainers covering a discrete set of learning rates $[0.1, 0.01, 0.001, 0.0001, 0.00001]$. Each one is trained with $1000000$ steps.
    \item \textit{GP-Bandit} \cite{gpbandit}: Gaussian Process Bandit with 5 parallel trainers covering a continuous range of learning rate $[10^{-5}, 10^{-1}]$ with logarithm scale. Each trial is trained with $10000$ steps.
    \item \textit{CMA-ES} \cite{CMAES}: A method based on covariance matrix adaptation evolution strategy with 5 parallel trainers covering a continuous range of learning rate $[10^{-5}, 10^{-1}]$ with logarithm scale. Each trial is trained with $10000$ steps. However, this method does not utilize warm-starting like PBT. 
    \item \textit{PBT-5x5}: 5 parallel trainers for a population of size 5, covering a continuous range of learning rate $[10^{-5}, 10^{-1}]$ with logarithm scale. Each trial is trained with $1000$ steps.
    \item \textit{PBT-5x20}: 5 parallel trainers using the budget mode simulating a population of size 20, covering a continuous range of learning rate $[10^{-5}, 10^{-1}]$ with logarithm scale. Each trial is trained with $1000$ steps.
\end{itemize}

\subsection{Convergence}

\subsubsection{Resource vs. accuracy.} 
\begin{figure}[!tb]
\centering
\includegraphics[width=\linewidth]{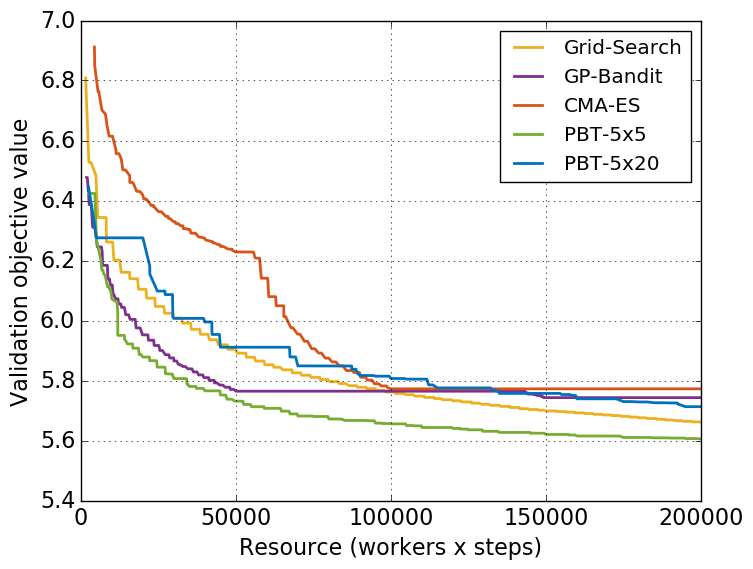}
\caption{Objective value vs. Resource: Resources are defined as the number of workers (each uses 18 GPUs) times the number of training steps performed on each worker. All methods use 5 parallel workers. Lower objective values are better.}
\label{fig:fair-compare}
\end{figure}

To perform a  fair comparison among different methods, we plot the objective values on different computation resources (total number of workers times the number of steps per worker). Please note each worker contains a trainer using 17 GPUs and an evaluator using another single GPU which adds up to a total of 18 GPUs. We did not use the actual time since there is variability in the time consumption due to factors such as disk and CPU congestion. We show the comparison in Figure \ref{fig:fair-compare}. Interestingly, PBT-5x5 outperforms all other methods including PBT-5x20. The reason why PBT-5x5 converges faster than PBT-5x20 is because using 5 workers to simulate a population size of 20 slows down the progress of generation by a factor of around 4. So the actual number of model training steps given the same resource limit is much lower in PBT-5x20. Please see Section \ref{sec:perf} for more details on the performance. 

\subsubsection{Continuing the training.} 
\begin{figure}[!tb]
\centering
\includegraphics[width=\linewidth]{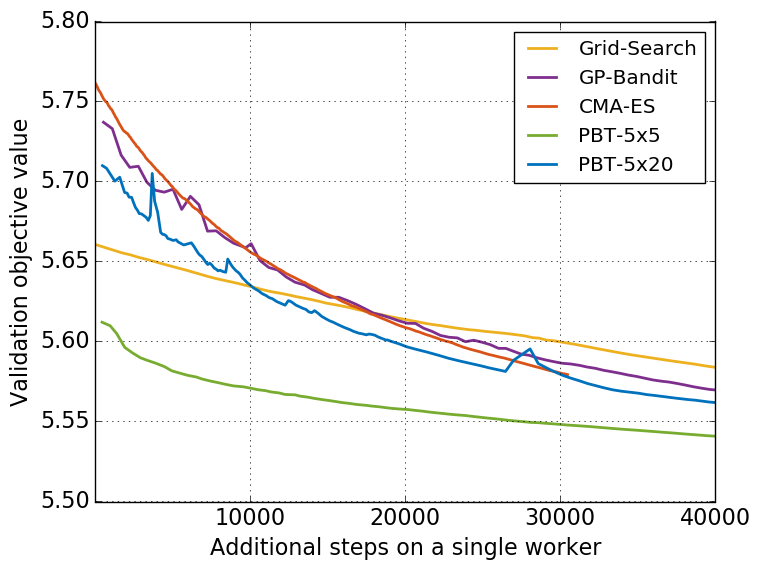}
\caption{Continue training on a single worker after 200000 resources exhausted, starting with the best checkpoint and its corresponding hyperparameters. Lower objective values are better.}
\label{fig:continuetrain}
\end{figure}

One often uses a fixed computational budget to perform hyperparameter tuning and pick the best set of hyperparameters to continue training the model. We show in this section that PBT can also be effective in a similar fashion, \textit{i.e.}, performing an initial joint optimization with a dynamic schedule of parameters and extend the training with the last (or best) set of hyperparameters. Specifically, we continue the training for all methods by taking the best checkpoint at resource 200K from Figure \ref{fig:fair-compare} and its corresponding hyperparameters. The best found learning rate in Grid search is $0.0001$, GP-Bandit $0.000329653$ and CMA-ES $0.000266482$. The learning rate at the best checkpoint in PBT-5x5 is $0.000114267$ and in PBT-5x20 is $0.000293463$. The comparison is shown in Figure \ref{fig:continuetrain}. All methods are trained using a single worker so they are essentially still utilizing the same computation budget. It is interesting to observe that although grid search outperforms Bandit, CMA-ES and PBT-5x20 within the initial 200k resource, its objective values quickly lift the highest during the continual training stage. PBT-5x5 consistently outperforms all other methods while PBT-5x20 catches up to second place.

\begin{figure}[!tb]
\centering
\includegraphics[width=0.95\linewidth]{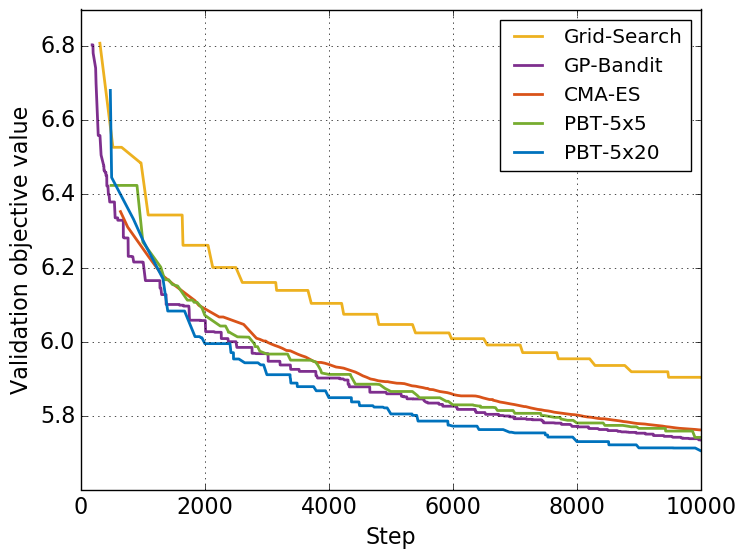}
\caption{Objective Value vs. Training Step: PBT with 20 population size outperforms all other methods. PBT with 5 population size performs in the second place, which shows that bigger population benefits the model accuracy. Lower values of the objective are better.}
\label{fig:step-obj}
\end{figure}

\subsubsection{Step vs. accuracy.} One may also want to reuse the discovered hyperparameters to re-train their models on a same or different dataset. The computation budget of re-training is solely dependent on the training steps. So we also plot the minimum objective values found within a certain number of training steps for all methods. The comparison is shown in Figure \ref{fig:step-obj}. We can see that PBT-5x20 outperforms all other methods including PBT-5x5. The reason why 20 population performs better than 5 population is because a larger population can achieve a higher coverage in the hyperparameter search space. PBT-5x5 is comparable to GP-Bandit while CMA-ES performs slightly worse. Grid search is significantly worse than all other approaches and it fails to find a better hyperparameter. However, it can achieves a larger number of training steps within a fixed budget. That is why grid search performs well in Figure \ref{fig:fair-compare}. 

The above results suggest that (1) given a fixed computation budget, a user should use a small population in order to see fast training progress for one-time training scenarios such as model development and debugging; (2) when the initial training budget is flexible and especially when a user expects to re-train the model in future occasions, one should consider using a large population to find an efficient hyperparameter schedule that could be reused.


\subsection{Dynamic hyperparameters}

\begin{figure}[!tb]
\centering
\includegraphics[width=0.95\linewidth]{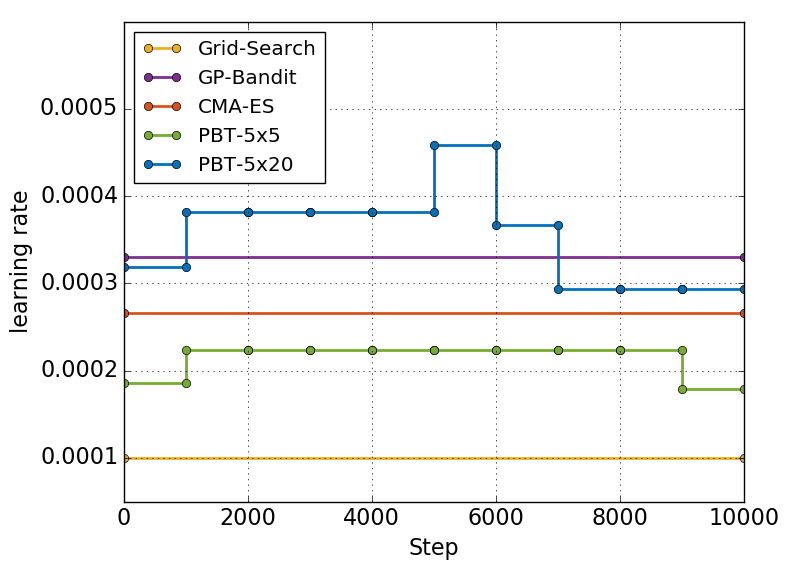}
\caption{Learning rate schedules found by different approaches.}
\label{fig:pbt-5x20-lr}
\end{figure}

An important property of population based training is its ability to discover a dynamic set of hyperparameters and train the model with such dynamic schedule at the same time. We extract the best learning rate schedule discovered by all methods for the first $10000$ steps, shown in Figure \ref{fig:pbt-5x20-lr}. We notice that PBT-5x20 finds a more dynamic schedule than PBT-5x5 which implies the fact that large population has better coverage in the hyperparameter search space. Also interestingly, the learning rate in PBT-5x20 went up to $0.000459$ in the first 6 generations ($5000$ steps) and then gradually decreated to $0.00029$. The phenomenon in the first 6 generations is similar to the known ``learning rate warm-up'', also used in training the Transformer network \cite{transformer}.

\begin{figure}[!tb]
\centering
\includegraphics[width=\linewidth]{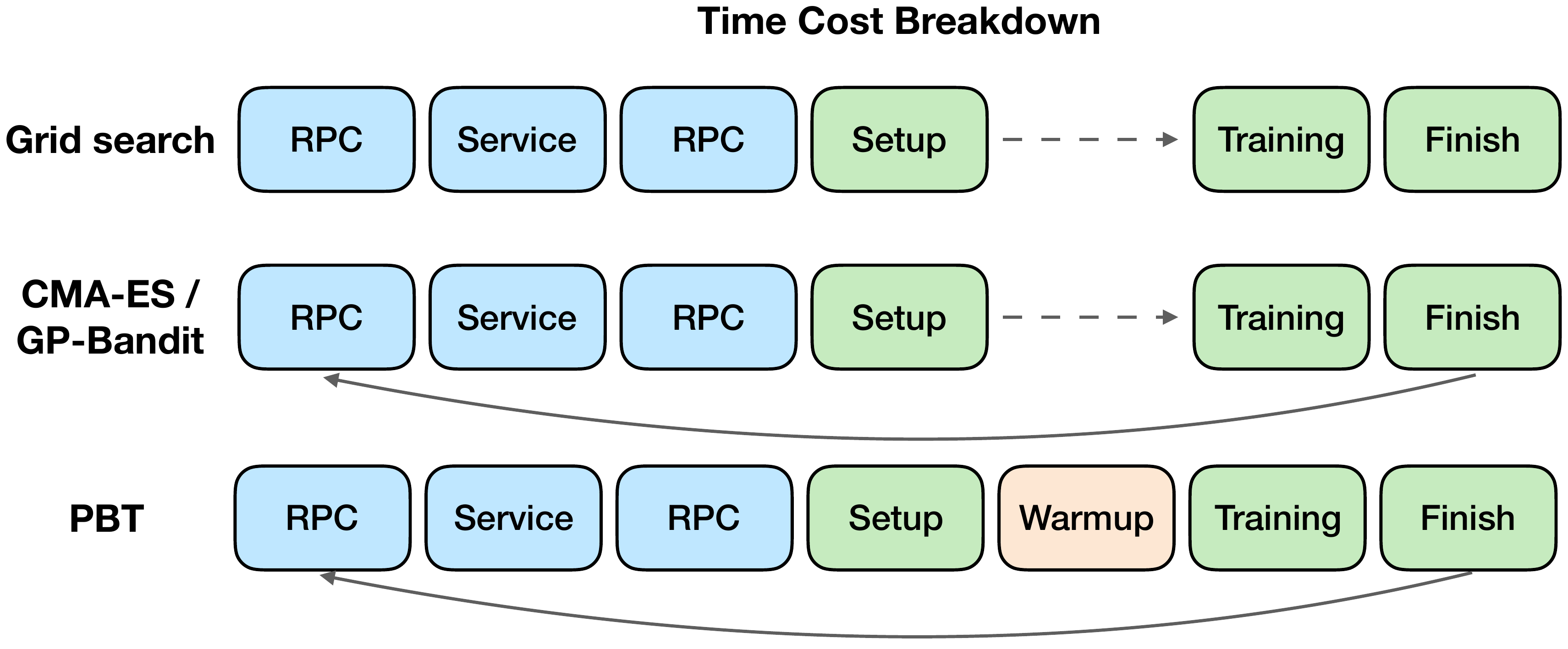}
\caption{Time cost breakdown for different methods.}
\label{fig:breakdown}
\end{figure}

\subsection{Performance}\label{sec:perf}
We analyze the performance for different methods in this section. We first show in Figure \ref{fig:breakdown} an illustration about the processes that every method needs to go through. 

\subsubsection{Computation breakdown.} The grid search approach needs (1) the worker to do a Remote Procedure Call (RPC) to the server, (2) the service to pick one discrete hyperparameter, (3) the service to return the hyperparameter to the worker through RPC, (4) the worker to setup training, (5) the worker to perform training and (6) the last finish-up procedure.

The GP-Bandit, CMA-ES and PBT methods need an additional process which returns the evaluation measurements to the service through RPC. And the PBT approach needs one more procedure which is the worker to warm start from a given checkpoint specified by the PBT service. So essentially, the PBT service has little additional computation on the worker side.

\begin{figure}[!tb]
\centering
\includegraphics[width=0.9\linewidth]{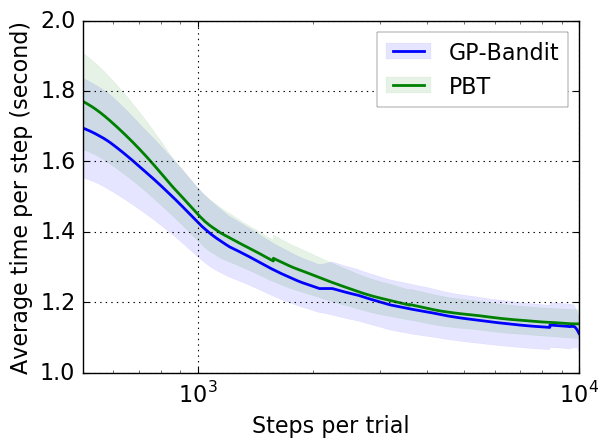}
\caption{The average time (seconds) per step varies when the number of steps per trial increases. PBT is slightly more expensive than GP-Bandit at the same number of steps ($+0.023s$ @ 1K and $+0.028s$ @10K), probably due to the extra warm-starting. The shaded area represents the $95\%$ confidence interval.}
\label{fig:avg_steptime}
\end{figure}

\subsubsection{Overhead.} It is important to notice that the PBT method uses $1000$ steps for each trial, which can produce overhead in the other components. So we compare the performance using the averaged time cost per step. The result is shown in Figure \ref{fig:avg_steptime}. The figure shows that the per-step time decreases as the number of steps of a trial increases. This implies that the initialization of a trial is the most expensive part. We found that PBT is slightly more costly (roughly $+0.025$ seconds) than GP-Bandit in every configuration. This is probably due to the additional warm starting at the beginning.

\subsection{Sensitivity}\label{sec:sensitivity}
Another important aspect of model training is its sensitivity to randomization. The same model can be trained multiple times. A stable performance is often desirable. We compare PBT with random search on their performance stability across different runs. Random search can also be seen as a special case of PBT where the number of training steps in each trial is infinite.

\begin{figure}[!tb]
\centering
\includegraphics[width=\linewidth]{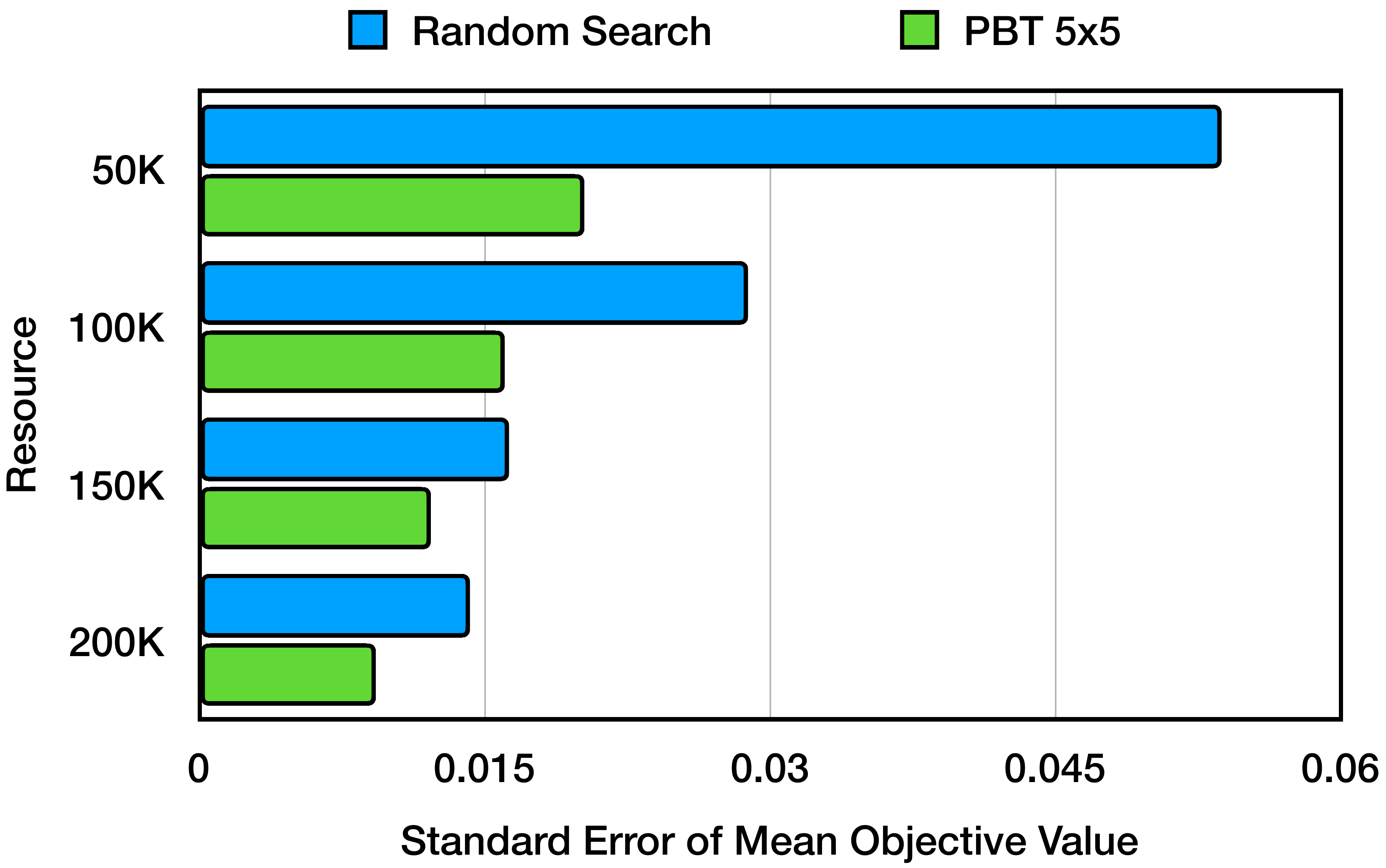}
\caption{Sensitivity analysis: comparing the standard errors of the mean objective between random search and PBT. The results are computed over 5 runs for each method. Lower value means less sensitivity.}
\label{fig:sensitivity}
\end{figure}

Figure \ref{fig:sensitivity} shows the standard errors of the mean (SEM) objective values at different resource level. The results are computed over five runs of either methods. The figure suggests that PBT is consistently more stable than random search across different runs.

\subsection{Scalability}
\begin{figure}[!tb]
\centering
\includegraphics[width=0.95\linewidth]{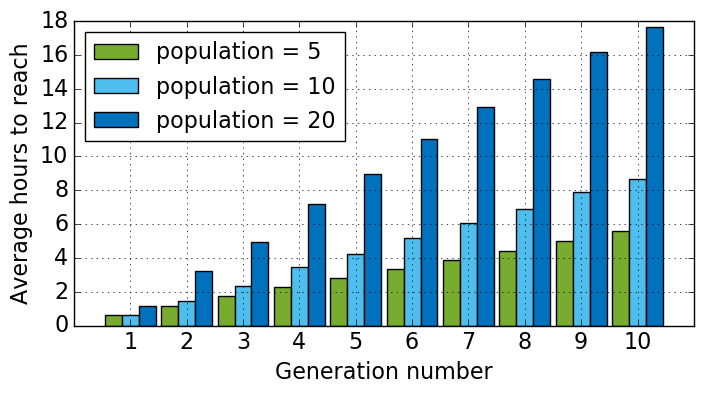}
\caption{Population size vs. the average time (hours) to reach a certain generation in the budget mode of population based training. The number of real workers for both is 5.}
\label{fig:popsize-vs-gen}
\end{figure}

\begin{figure}[!tb]
\centering
\subfigure[\label{fig:popsize-vs-gen-per-hour}]{
\includegraphics[width=0.45\linewidth]{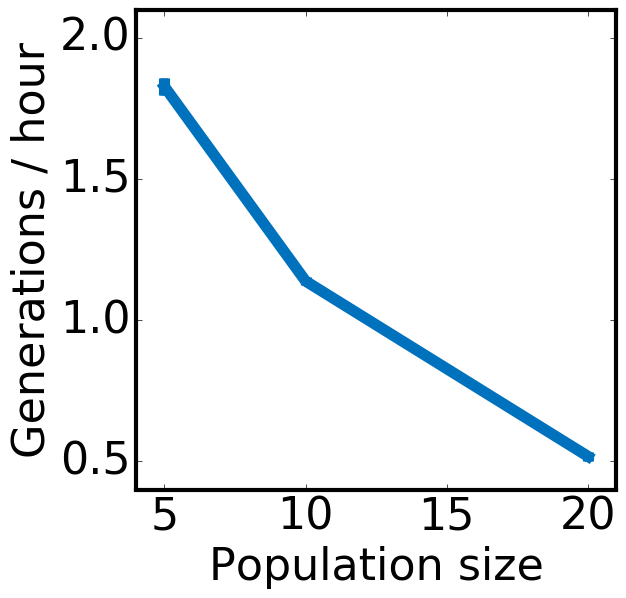}~~~}
\subfigure[\label{fig:worker-vs-gen-per-hour}]{
\includegraphics[width=0.45\linewidth]{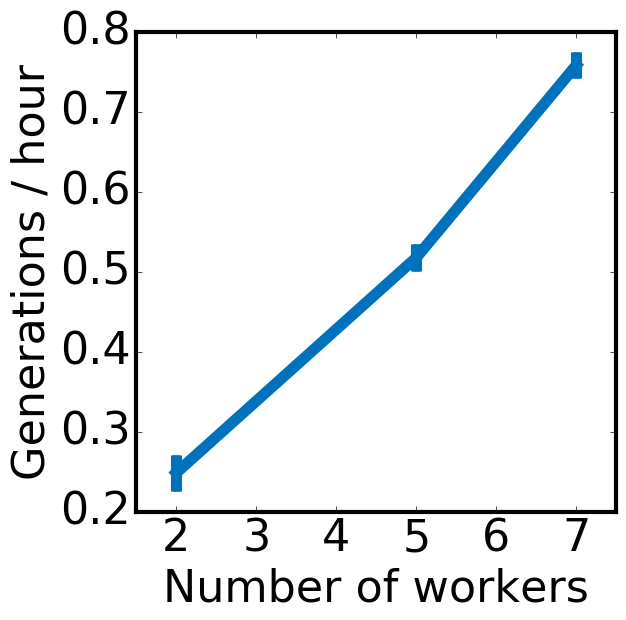}~~~}
\caption{Scalability: The number of progressed generations per hour vs (a) varying population size or (b) varying worker number of population based training. In (a), the number of (real) workers is 5. In (b), the population size is 20. The error represents the $95\%$ confidence interval.}
\end{figure}

As we mentioned above, the population size can affect accuracy and performance in PBT. We conducted a study on how the performance varies with a different population size. We show in Figure \ref{fig:popsize-vs-gen} the average time (hours) for PBT to reach a certain generation number. The result shows that PBT with population size 20 costs roughly 3x the computation of PBT with population size 5 (theoretically it should be 4x but there are some variation in the real cluster environment). We further plot the number of generations progressed per hour under different population sizes in Figure \ref{fig:popsize-vs-gen-per-hour} and different number of workers in Figure \ref{fig:worker-vs-gen-per-hour}. Both results suggest a nearly linear scalability of our system.

\subsection{Opponent Selection Strategy}\label{sec:opponent}
\begin{figure}[!tb]
\centering
\includegraphics[width=\linewidth]{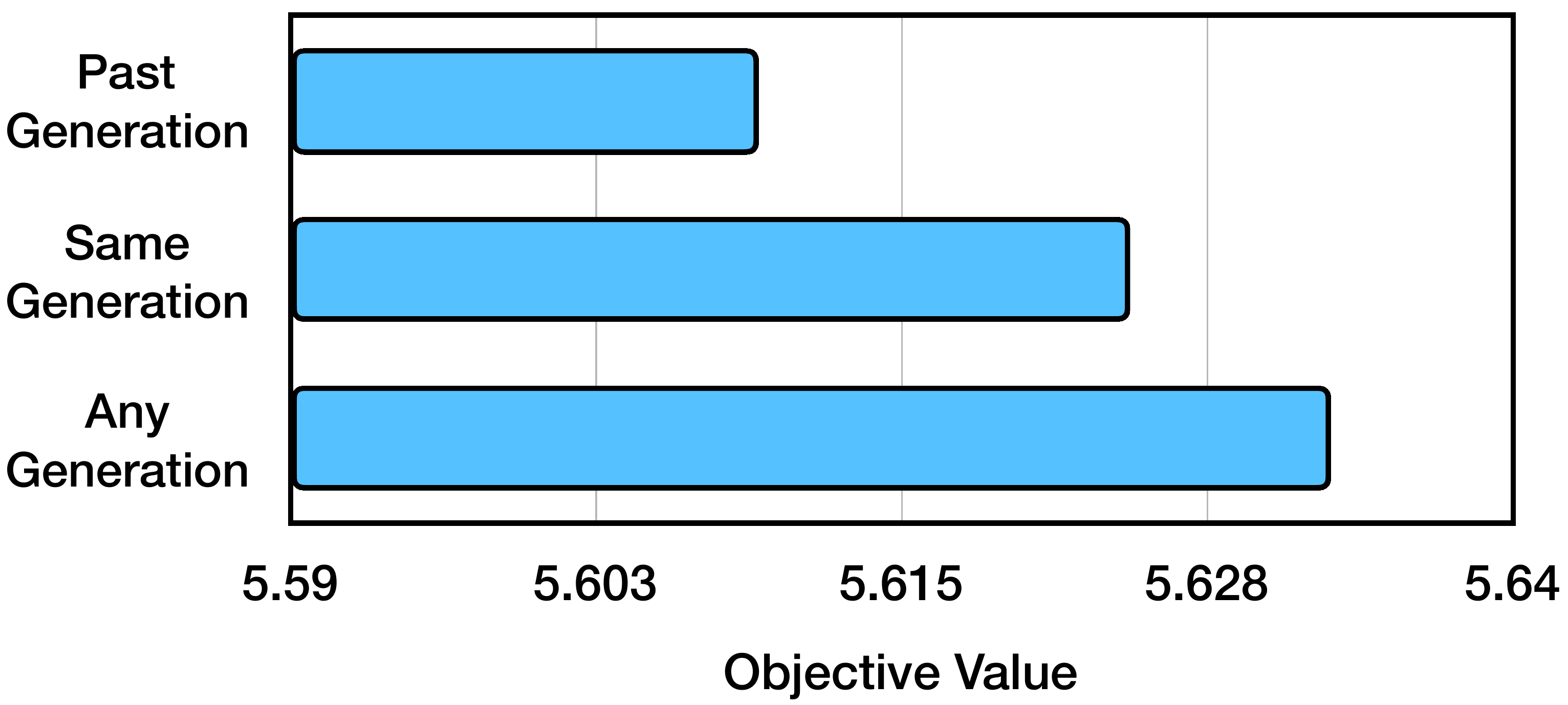}
\caption{Comparison of opponent selection strategies using the objective values at resource 200K. Lower is better.}
\label{fig:opponent-selection}
\end{figure}

To justify the design choice in the proposed evolution framework, we perform a comparison among three opponent selection strategies. The objective values at resource 200K are compared in Figure \ref{fig:opponent-selection}. ``Past generation'' is our design choice which means competing with trials in the earlier generations and same generation. ``Same generation'' means only competing with trials in the same generation. ``Any generation'' allows a trial to compete with future generations. The result suggests that ``past generation'' results in the best performance. A possible reason is because this method relieve the effect of the speed differences in different workers.

\section{Related Work}

\subsection{Hyperparameter Optimization}

Hyperparameter optimization is a critical step in most machine learning systems \cite{NIPS2011_hyper}. Over the past years, there have been a variety of algorithms proposed to automatically tune hyperparameters of a machine learning model including Grid search, Random search \cite{randomsearch}, Bayesian optimization \cite{Snoek:bayesian}, Gradient-based optimization \cite{gradient-hparam}, Bandits \cite{Li2017HyperbandAN} and Evolutionary algorithms \cite{CMAES}.

Population based training \cite{max-pbt-2017} is the core algorithm of our service. Our implementation is different from the original paper in that the workers are not allowed to access the measurements of other workers and mutation decisions are not made inside the worker process. The advantage of our black-box design is that it allows a user to train using PBT with minimal infrastructure changes.

Google Vizier \cite{vizier} is the most related system which is a hyperparameter tuning service. A user receives trial information from the Vizier service, training with the specified hyperparameters and returning the measurements to the service. PBT service inherits the service design from Vizier while exntending to require the trainers to always warm start from a given checkpoint path.

In addition, the PBT service is a distributed model training service whose purpose is broader than just hyperparameter tuning. PBT is a joint learning process that combines both hyperparameter search and model training into one single loop. So the outcome of PBT is not only a hyperparameter schedule but also a set of high performing models. A user can either choose the existing best model checkpoint for serving or extract the best hyperparameter schedule to perform other training procedures.




\subsection{AutoML via Asynchronous Evolution}
Evolution algorithms, naturally distributed, have been widely applied to a variety of applications at scale. For hyperparameter tuning in machine learning models, the covariance matrix adaptation evolution strategy (CMA-ES) has been popular \cite{FRIEDRICHS2005107, CMAES}.

Recently, large scale distributed evolution has proven effective in the neural architecture search problem. \citet{pmlr-v70-real17a} is probably the first to apply the evolution framework into large scale neural architecture search. More recently, \citet{amoebanet} introduced a simple regularization to the asynchronous evolution framework, which removes population members according to their ages and produces appealing results in searching image classifier architectures.

Most of the existing evolution based machine learning systems do not full exploit the idea of warm-starting which is a core component in the PBT system design. Warm-starting a model training allows for the transfer of existing knowledge gained in previously trained models and enables efficient hyperparameter search.

\section{Conclusion}
We presented a black-box service framework for Population Based Training. The proposed service design allows clients to train their models using PBT with minimal infrastructure effort, \textit{i.e.}, the only requirement of applying PBT to an existing training framework is to warm start the model from a checkpoint and report the measure back to the PBT server. We discussed several useful features in our PBT system such as training replay, training recovery, garbage collection, and budget mode. We conducted a case study of our system in WaveNet human speech synthesis and demonstrated that our PBT system produces superior accuracy and performance compared to other popular hyperparameter tuning methods. Moreover, the PBT system is able to directly train a model using the discovered dynamic set of hyperparameters while traditional methods can only tune static parameters. In addition, we show that the proposed PBT framework is feasible for large scale deep neural network training.

\section*{Acknowledgement}
The authors would like to thank Alexander Vostrikov, Ali Razavi, Andy Bodart, Ben Coppin, Daniel Golovin, Daniel Visentin, Eddie Kessler, Eli Bixby, Gabriel Doliner, Harish Chandran, Joyce Chen, Karen Simonyan, Mariam Doliashvili, Matthieu Devin, Maxim Krikun, Michelle Gong, Oriol Vinyals, Salem Haykal, Sibon Li, Simon Osindero, Todd Hester, Tom Walters, Vivek Ramavajjala, Zhe Zhao, Zora Tang and many other colleagues in Alphabet for their meaningful discussions and contributions.

\bibliographystyle{ACM-Reference-Format}
\bibliography{pbtpaper}

\end{document}